\documentclass{article}

\usepackage{microtype}
\usepackage{graphicx}
\usepackage{booktabs}
\usepackage{hyperref}

\usepackage[accepted]{icml2026}
\usepackage{amsmath}
\usepackage{amssymb}
\usepackage[capitalize,noabbrev]{cleveref}
\usepackage{xspace}
\usepackage{orcidlink}
\usepackage{fontawesome5}
\makeatletter
\newcommand{\github}[1]{\href{#1}{\faGithub}}

\newcommand{\clax}{\textsc{clax}\xspace}
\newcommand{\claxpt}{\textsc{clax-pt}\xspace}

\newcommand{\classpt}{\textsc{class-pt}\xspace}
\newcommand{\jax}{\textsc{jax}\xspace}

\icmltitlerunning{Physics Is All You Need? A Case Study}

\begin{document}

\twocolumn[
  \icmltitle{Physics Is All You Need? A Case Study in Physicist-Supervised \\
    AI Development of Scientific Software}

  \icmlsetsymbol{equal}{*}

  \begin{icmlauthorlist}
    \icmlauthor{Nhat-Minh Nguyen\,\orcidlink{0000-0002-2542-7233}}{kavli,cd3,icise}
  \end{icmlauthorlist}

  \icmlaffiliation{kavli}{Kavli IPMU (WPI), UTIAS, The University of Tokyo, 5-1-5 Kashiwanoha, Kashiwa, Chiba 277-8583, Japan}
  \icmlaffiliation{cd3}{Center for Data-Driven Discovery, Kavli IPMU (WPI), UTIAS, The University of Tokyo, Kashiwa, Chiba 277-8583, Japan}
  \icmlaffiliation{icise}{Institute For Interdisciplinary Research in Science and Education, ICISE, Quy Nhon, 55121, Vietnam}

  \icmlcorrespondingauthor{Nhat-Minh Nguyen}{nhat.minh.nguyen@ipmu.jp}

  \icmlkeywords{AI-assisted development, scientific software, human-AI collaboration, supervision protocols}

  \vskip 0.3in
]

\printAffiliationsAndNotice{}

\begin{abstract}
Are AI agents tools, co-authors, or researchers? We present a quantified case study ($N=1$): a physicist supervising an AI coding agent (Claude Code, Sonnet and Opus models) over 12 work days and 57 sessions to build a differentiable one-loop perturbation theory (a next-to-leading-order calculation for predicting galaxy clustering) module in JAX, \claxpt\href{https://github.com/MinhMPA/clax-pt}{\faGithub} (${\sim}2{,}100$ lines, validated to ${\lesssim}1\%$ accuracy against the established C reference \classpt). We documented 15 supervision events during the v0.1.0 development window and classified each by intervention level (Appendix~\ref{app:bug-table}).

The agent resolved ten autonomously (convention errors, algorithm transcription, numerical coefficients) by iterating against oracle test suites. Two more were accelerated by the physicist spotting magnitude discrepancies invisible to shape-based comparisons. The three it could not---all evaded oracle detection---share a common property: the agent treated symptom reduction as equivalent to root-cause resolution. It spent 33 of the 57 sessions adjusting coefficients within a code architecture that could not represent the target physics, and could not re-evaluate its choice of \classpt branch even when the physicist explicitly prompted reconsideration; only an injected physics concept (anisotropic BAO damping) triggered the redesign. Separately, the agent committed a calibrated scalar correction that passed all oracle tests, but the value corresponded to no quantity in the reference theory and would produce wrong predictions at any other cosmology.

The fudge factor was caught and replaced within the same session. Three supervision practices, developed iteratively, proved critical for catching what oracle tests missed: testing at diverse parameter points beyond the fiducial calibration; shared changelogs that surfaced stalled exploration across sessions; and an explicit rule against unphysical numerical patches. In this case study, the design of these supervision protocols, not model capability, was the primary factor in whether the agent's output was trustworthy. Closing the gap we observed would require agents that can propose architectural alternatives rather than optimize within a given structure, and distinguish predictive adequacy from explanatory correctness. The agent in this case study did not exhibit these capabilities, and they are not obviously addressed by scaling alone.
\end{abstract}

\section{Introduction}
\label{sec:intro}

Most empirical evidence on AI coding agents in science comes from one of two regimes: standardized coding benchmarks and large-scale software projects where automated testing fully captures correctness~\citep{carlini2026}, or fully autonomous multi-agent systems that generate manuscripts without sustained human oversight~\citep{villaescusa2025}. Neither captures the reality of scientific software development, where correctness is defined by agreement with physical law rather than by compilation or test passage, and where physicists work with AI agents to produce and validate a scientific software.

We present a detailed case study of exactly this scenario. Over 12 work days and 57 sessions, a physicist supervised an AI coding agent (Claude Code, Sonnet and Opus models) to build \claxpt\href{https://github.com/MinhMPA/clax-pt}{\faGithub}, a differentiable one-loop perturbation theory module (computing next-to-leading-order predictions for galaxy clustering) in \jax. The module computes nine output power spectra validated to ${\lesssim}1\%$ accuracy against the established reference code \classpt~\citep{chudaykin2021}. The resulting implementation comprises approximately 2,100 lines of code. The contribution of this paper is not the code itself (described in a companion paper) but the empirical supervision record: what the human did, what the agent did, and where the boundary between their roles became load-bearing.

Our central claim is that the supervision protocol---not model capability---was the primary factor determining whether the agent's output was trustworthy scientific software. The agent autonomously resolved 10 of the 15 documented issues by iterating against oracle test suites. Two more were accelerated by the physicist's domain knowledge input. The three it could not resolve were invisible to oracle testing and required human physics judgment to diagnose. These three bugs consumed a disproportionate fraction of the project: 33 of 57 sessions were spent within a fundamentally wrong code architecture that the agent could not recognize as wrong because it passed local consistency checks. The physicist's interventions---an architectural redesign and the rejection of a physically unmotivated numerical patch---were the decisive acts that produced correct code. Both share a common cognitive structure. The physicist assessed not whether the code produced right numbers, but whether it produced them \emph{for the right reasons}.

\section{The Project}
\label{sec:project}

\subsection{What Was Built}
\label{sec:what}

We compute predictions for the gravitational clustering of galaxies, the three-dimensional distribution mapped by galaxy redshift surveys. The calculation involves loop integrals (analogous to next-to-leading-order corrections in quantum field theory) that are expensive to evaluate and sensitive to subtle physical effects: bulk motions of matter and oscillation features imprinted by sound waves in the early universe. Getting these effects wrong by even 1\% can bias the cosmological parameters (the parameters of the standard model of cosmology) inferred from the galaxy power spectrum.

One-loop perturbation theory (the next-to-leading-order correction beyond the linear approximation) extends the linear matter power spectrum to mildly nonlinear scales by computing correction terms from second- and third-order density fields, producing predictions for galaxy clustering that are essential for extracting cosmological parameters from spectroscopic surveys~\citep{Ivanov:2019pdj, DAmico:2019fhj}. \claxpt implements this calculation in \jax: it takes a linear power spectrum and cosmological parameters as input, evaluates the tree-level and one-loop terms (the latter via FFTLog decomposition~\citep{simonovic2018, fang2017}), applies infrared (IR) resummation~\citep{senatore2015, vlah2015}, and adds ultraviolet (UV) counterterms that absorb sensitivity to small-scale physics outside the perturbative regime~\citep{Baumann:2010tm, Carrasco:2012cv, Carroll:2013oxa}. IR resummation corrects for large-scale bulk flows which, if unaccounted for, artificially smear the baryon acoustic oscillation (BAO) feature, the standard-ruler imprint of primordial sound waves. The code returns nine validated output spectra (\cref{tab:accuracy}). The implementation is approximately 2,100 lines of \jax code, end-to-end differentiable, and validated to ${\lesssim}1\%$ accuracy against \classpt~\citep{chudaykin2021}. Technical details are described in a companion paper.

\begin{table}[t]
\caption{Accuracy of \claxpt vs.\ \classpt at Planck 2018 fiducial cosmology, $z=0.38$, $k < 0.3\;h/$Mpc. Hexadecapole uses $|\Delta|/\max(|\mathrm{ref}|)$ due to zero crossings.}
\label{tab:accuracy}
\centering
\footnotesize
\setlength{\tabcolsep}{4pt}
\begin{tabular}{@{}lcc@{}}
\toprule
Spectrum & Max error & Mean error \\
\midrule
$P_{mm}, P_{gg}, P_{gm}$ (real-space) & 0.31\% & 0.04\% \\
Monopole ($\ell=0$) & 0.59\% & 0.40\% \\
Quadrupole ($\ell=2$) & 0.89\% & 0.50\% \\
Hexadecapole ($\ell=4$) & 1.43\% & 0.37\% \\
\bottomrule
\end{tabular}
\end{table}

\subsection{The Supervision Protocol}
\label{sec:protocol}

Before any code was written, the physicist established a supervision protocol adapted from the methodology of Anthropic's C-compiler agent project~\citep{carlini2026}, which used 16 parallel agents over 2,000 sessions to build a compiler that successfully builds the Linux kernel. Four infrastructure elements structured every session:

\begin{enumerate}
\item \textbf{\classpt as oracle.} Every function was tested against reference data generated by the established C implementation. Tests were written before code---the agent knew what the correct output looked like before attempting to produce it.

\item \textbf{CHANGELOG as shared memory.} Because each agent session starts with no memory of previous sessions, a structured log recorded what was attempted, what failed, and what succeeded. This prevented later sessions from re-exploring dead ends (e.g., a discrete sine transform (DST) grid bug that was solved on day 3 was never re-investigated).

\item \textbf{A \texttt{--fast} flag for context-window hygiene.} Tests printed at most 10 lines on success and 20 on failure. Verbose diagnostics were written to log files. This discipline (drawn directly from \citet{carlini2026}) prevented the agent's finite context window from being consumed by noise.

\item \textbf{Parallel agent sessions via git worktrees.} Multiple sessions explored competing hypotheses simultaneously. When a bug had multiple plausible causes, the physicist spawned parallel investigations rather than serializing them.
\end{enumerate}

Two rules proved critical beyond this infrastructure. First, ``no fudge factors'': if a test failed at 0.2\% error, there was a real bug to find---a multiplicative correction tuned to make the test pass would be rejected at review. Second, ``test at multiple parameter points'': the oracle compared not just at the fiducial Planck 2018 cosmology but at varied cosmological parameters, to prevent solutions that were calibrated to a single point. The supervision protocol---not individual code edits---was the physicist's primary contribution to the project.

\section{Bug Taxonomy: The Autonomy Spectrum}
\label{sec:bugs}

\begin{figure}[t]
\centering
\includegraphics[width=\columnwidth]{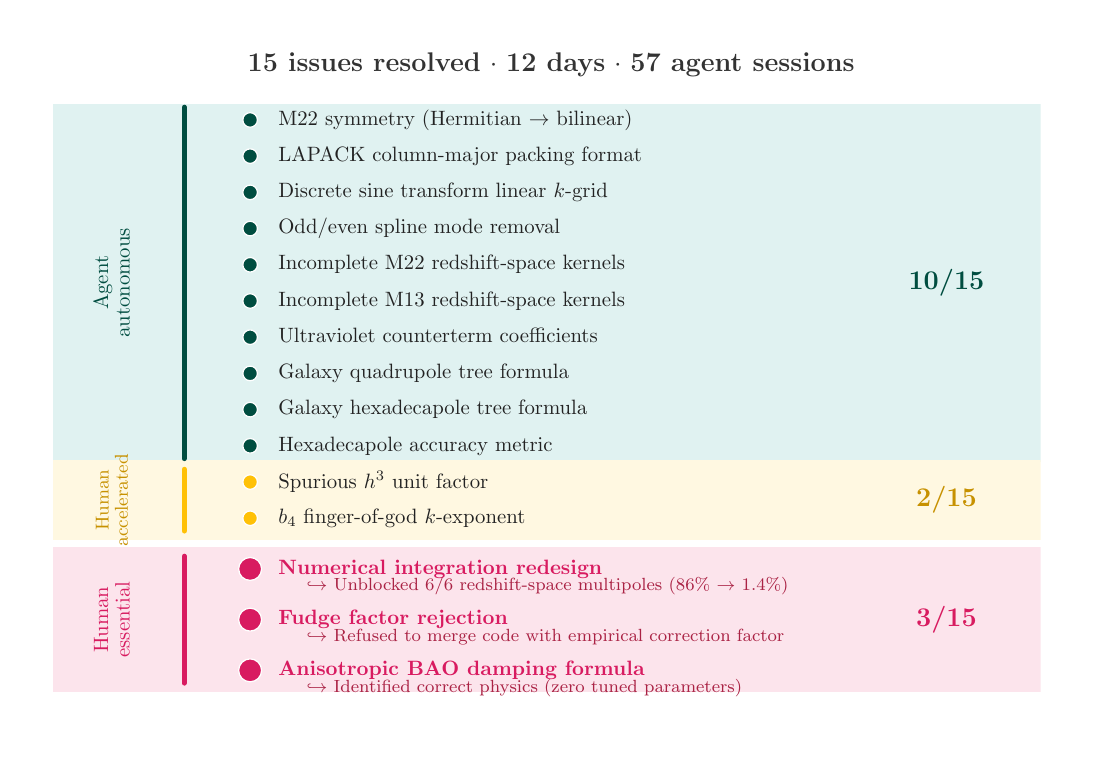}
\caption{Issue taxonomy for the \claxpt v0.1.0 development. Of 15 documented supervision events, 10 were resolved autonomously by the agent iterating against oracle tests; 2 were accelerated by the physicist's domain knowledge (unit-magnitude and dimensional discrepancies invisible to shape-based comparisons); 3 required essential human judgment (an architectural redesign, a calibration-patch rejection, and identification of the correct anisotropic damping formula). Bug-level classification with confidence flags and independent cross-check provenance is provided in Appendix~\ref{app:bug-table}.}
\label{fig:taxonomy}
\end{figure}

We documented 15 issues over 57 agent sessions and classified each by whether autonomous resolution was possible (\cref{fig:taxonomy}). The taxonomy that emerged is not a clean binary---it is a spectrum ranging from bugs the agent resolved in minutes to bugs that resisted 33 sessions of autonomous effort and required human physics judgment to diagnose.

\subsection{Autonomous and Human-Accelerated Issues}
\label{sec:autonomous}

The agent resolved 10 bugs without human intervention by iterating against the oracle test suite. Two more (the $h^3$ unit factor and $b_4$ $k$-exponent) were accelerated by the physicist spotting magnitude discrepancies invisible to shape-based comparisons. These fell into three categories. First, convention and unit errors: the FFTLog matrix (used to decompose loop integrals) $M_{22}$ required Hermitian rather than symmetric packing (bug 1), LAPACK column-major ordering differed from the row-major layout the agent initially produced (bug 2), and a spurious $h^3$ factor multiplied all bias functions (bug 5). Second, algorithm implementation errors: a DST grid used linear rather than logarithmic spacing (bug 3), odd/even spline interpolation for separating the BAO oscillation feature used the wrong parity assignment (bug 4), and three redshift-space distortion (RSD) kernel matrices were incomplete (bugs 7--9). Third, numerical coefficient errors in UV counterterms required matching specific rational prefactors from the 14,000-line C reference source.

In all these cases, the pattern was the same: the oracle test produced a clear numerical discrepancy, the agent compared intermediate quantities against reference data to localize the error, and the fix was a direct transcription correction.

Beyond bug-fixing, the agent performed codebase reconnaissance the supervisor had not directed. During the first sessions of bring-up, the agent mapped the structure of the \classpt reference source on its own, identifying that the effective field theory calculation lives in two parallel code paths with different treatments of the redshift-space integrals, without being told to look for this distinction or given a guide to the source layout. This survey was complete enough that, once the physicist later supplied the conceptual hint about anisotropic damping (\cref{sec:wall}), the agent could re-target its implementation to the previously-surveyed second branch without further retrieval. The codebase scan, then, was not the limitation in this case; the limitation was the agent's inability to re-evaluate which of the two mapped branches its failing tests implicated.

\subsection{Case Study: The Redshift-Space Multipoles}
\label{sec:wall}

\begin{figure*}[t]
\centering
\includegraphics[width=\textwidth]{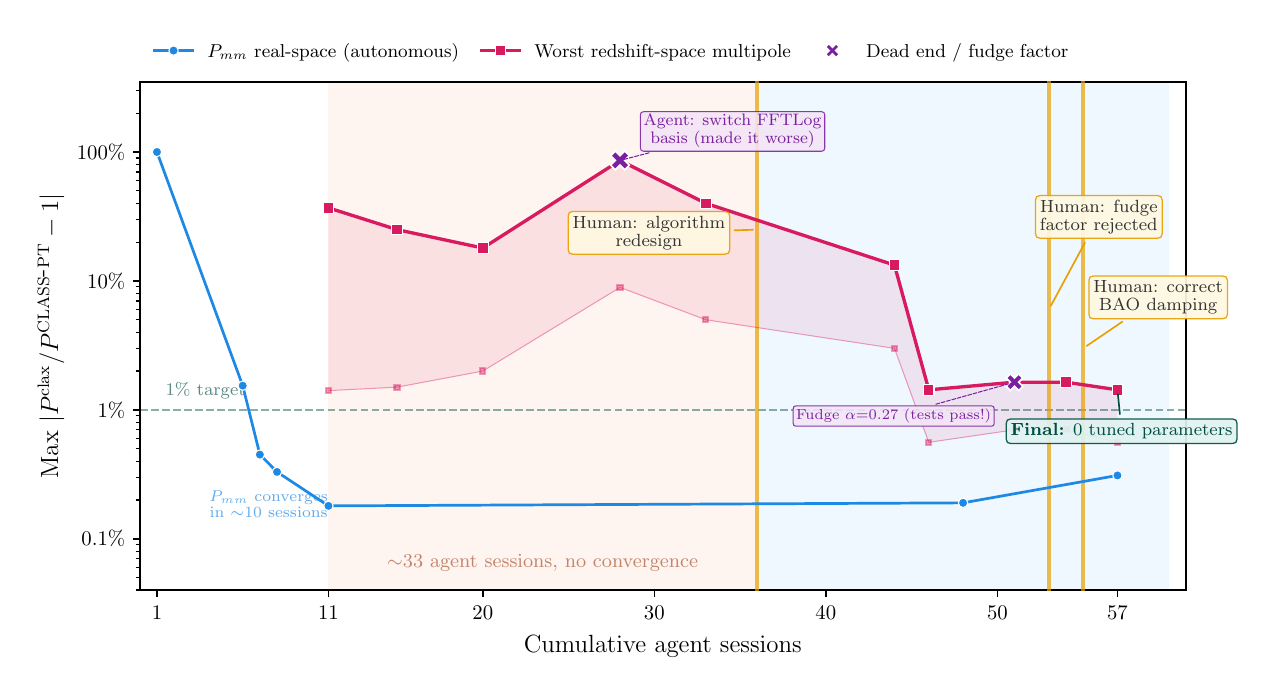}
\caption{Accuracy convergence over 57 agent sessions. Blue: real-space matter power spectrum (autonomous, converges in ${\sim}10$ sessions). Red: worst redshift-space multipole (stuck at $8$--$86\%$ for ${\sim}33$ sessions). Vertical gold lines mark human interventions. The agent's error plateaued because the code architecture was structurally incompatible with anisotropic BAO damping---no coefficient adjustment within that architecture could make all multipoles pass simultaneously. After the physicist directed a Gauss--Legendre quadrature redesign, errors dropped to the 1--2\% range. A subsequent fudge factor ($\alpha = 0.27$, purple cross) passed all tests but was rejected by the physicist as physically unmotivated. The correct formula required zero tuned parameters.}
\label{fig:wall}
\end{figure*}

\paragraph{The accuracy wall.}
After the first 24 sessions resolved bugs 1--9, all three real-space power spectra passed at sub-percent accuracy. The six RSD multipoles (monopole, quadrupole, and hexadecapole for both matter and galaxies) did not. Errors ranged from 8\% to 86\% depending on the multipole and varied erratically as the agent adjusted individual terms (\cref{fig:wall}).

Over the next 33 sessions (sessions 24--56 of 57 total), the agent attempted to fix the RSD multipoles within the existing code architecture. This architecture computed analytic Legendre projections of the one-loop integrands: for each multipole $\ell$ and each component (velocity-velocity, velocity-density, density-density), a pre-derived kernel matrix encoded the angular projection. The approach is correct when the infrared resummation factor (the exponential damping at BAO scales) is isotropic, meaning it depends only on wavenumber $k$ and not on the angle $\mu$ between the wavevector and the line of sight.

Over these 33 sessions, the agent worked coherently within this architecture: adjusting kernel coefficients, adding angular terms, swapping quadrature rules, and comparing matrix elements against the reference source term by term. Each fix improved one multipole while degrading another. The error surface was non-convex within this model: no combination of coefficient adjustments could make all six multipoles pass simultaneously, because the architecture itself was structurally incompatible with the physics.

The physicist diagnosed the structural issue by recognizing what the agent could not: the BAO damping in redshift space is \emph{anisotropic}. Peculiar velocities along the line of sight enhance the displacement field. The damping factor depends on the angle $\mu$ between the wavevector and the line of sight:
$\Sigma^2_\mathrm{tot}(\mu) = (1+f(2+f)\mu^2)\sigma^2_v + f^2\mu^2(\mu^2 - 1)\sigma^2_\mathrm{BAO}$,
where $f$ is the linear growth rate of structure (how fast density perturbations amplify over time). This $\mu$-dependent exponential cannot be absorbed into a polynomial kernel matrix---it couples to the RSD Kaiser factor $(1+f\mu^2)^2$ (the leading-order enhancement of clustering along the line of sight due to galaxy peculiar velocities) in a way that no finite set of pre-computed analytic projections can represent exactly.

The physicist proposed a redesign\footnote{Physicist-side authorship of this redesign is independently corroborated by a parallel Codex CLI session the day before the GL-redesign commit, in which the same author asked the Mathematica side of the analysis to parallelize the multipole computation over $\mu$ samples---the architectural move that landed in \claxpt next. See supervision log.}: abandon per-multipole analytic kernels entirely, assemble the full anisotropic power spectrum $P(k,\mu)$ at each of $N$ Gauss--Legendre quadrature nodes in $\mu$, and integrate numerically against Legendre polynomials to extract multipoles. The agent implemented this redesign in a single session. Errors for all six RSD multipoles dropped from 8--86\% to the 1--2\% range, with four of six passing sub-percent immediately. The remaining two quadrupoles required an additional correction.

Why could the agent not find this itself? The agent had autonomously surveyed both \classpt code paths during initial exploration: the simpler path (isotropic damping with analytic Legendre projections) and the more complex path (anisotropic damping handling geometric distortion corrections). It selected the simpler path as its implementation target---a reasonable choice given the test configuration did not require geometric distortion---and proceeded to implement it. What it could not do, across 33 sessions of unsuccessful coefficient adjustment, was \emph{re-evaluate} that selection: ask whether the path it had not chosen might be the relevant one for the failing tests. The physicist explicitly attempted process scaffolding (a generic, domain-free reconsider-the-architecture prompt: ``the current architecture may be the wrong frame; please reconsider whether your existing kernel-matrix structure can represent the target physics, rather than tuning coefficients within it''); the agent reaffirmed its design and continued coefficient adjustments. The architectural redesign was triggered only when the physicist supplied the relevant physics concept (anisotropic BAO damping in redshift space). Given that concept, the agent recognized the previously-surveyed second branch as the appropriate implementation target.

Codebase retrieval was already complete in this case, so we did not run a controlled ablation isolating the conceptual injection from a hypothetical navigational-pointer-only intervention. We therefore cannot rule out that aggressive retrieval over the full \classpt codebase, delivered to a different agent without our reconsider-the-architecture prompt, might have surfaced the anisotropic branch unprompted. Within this case, the gap was in generating the physics question that selects between mapped alternatives---not in scanning the codebase.

\paragraph{The fudge factor.}
After the Gauss--Legendre redesign, seven of nine spectra passed. Two quadrupole spectra ($P_{mm,\ell=2}$ and $P_{gg,\ell=2}$) failed at scales near the BAO peak with errors of ${\sim}1$--$2\%$. The agent traced the discrepancy to the resummed tree-level spectrum, where the BAO damping factor enters multiplied by a polynomial correction:
\[
P_\mathrm{tree}(k) = P_\mathrm{nw}(k) + e^{-\Sigma^2 k^2}(1 + \Sigma^2 k^2)\,P_\mathrm{w}(k),
\]
with $P_\mathrm{nw}$ and $P_\mathrm{w}$ the no-wiggle and wiggle components and $\Sigma^2$ the BAO damping scale.

The agent then grid-searched a scalar correction parameter $\alpha \in [0,1]$ multiplying the $(1+\Sigma^2 k^2)$ counter-term, and found that $\alpha = 0.27$ minimized the worst-case error to below 1\% across all nine spectra simultaneously. The agent committed this fix with a passing test suite. The ``no fudge factors'' rule had been part of the written supervision protocol from the start of the project. The agent did not flag $\alpha=0.27$ as a violation, framing $\alpha$ as a free parameter inside an existing physics formula rather than as a tuned correction. The literal rule was followed; the principle was missed.

This solution is wrong, despite passing every oracle test. The value $\alpha = 0.27$ has no physical derivation. It is not a parameter in \classpt, and not derived from any perturbation theory calculation. The EFT bias coefficients and UV counterterms of the same calculation are also free parameters, but each absorbs a specific physical effect within a controlled perturbative expansion; $\alpha$ has no such derivation. It is a numerical calibration to the fiducial Planck 2018 cosmology at a single redshift, and would produce predictions wrong at any other cosmology by an amount invisible to a test suite that only checks the fiducial point.

The physicist diagnosed the fudge factor by asking a question the agent could not formulate: ``what does $\alpha$ correspond to in the perturbation theory derivation, or did you just copy it from the \classpt source?'' The agent confirmed neither: $\alpha$ appears nowhere in the reference, and no derivation produces it. The physicist then pointed out that the tree-level spectrum $P_\mathrm{tree}$ (the building block of the one-loop calculation) should carry the same anisotropic damping $\Sigma^2_\mathrm{tot}(\mu)$ around the BAO scale that already appeared in the Gauss--Legendre loop, rather than the isotropic form the agent had inherited from the simpler \classpt branch. The fix was to move the tree-level computation inside the existing quadrature loop---three lines of code, replacing the committed fudge factor with the correct anisotropic formula. After this fix, all nine spectra passed with zero tuned parameters.

The fudge factor illustrates a failure mode specific to oracle-tested scientific software. A numerically adequate but physically meaningless solution passes all existing tests. The agent cannot distinguish between ``this works because it is correct'' and ``this works because it is calibrated to the test data.'' The physicist could, because the physicist knew that $\alpha = 0.27$ must correspond to \emph{something} in the theory---and when it corresponded to nothing, the solution was suspect regardless of its numerical performance.

The three human-essential interventions were thus: (i)~the Gauss--Legendre architectural redesign that unblocked all six redshift-space multipoles (\cref{sec:wall}), (ii)~the rejection of the fudge factor $\alpha=0.27$ as physically unmotivated, and (iii)~the identification of the correct anisotropic tree-level formula. Items (ii) and (iii) occurred in the same debugging episode but represent distinct cognitive acts: rejecting a wrong answer versus finding the right one.

\section{Lessons and Limitations}
\label{sec:lessons}

The case study suggests three patterns observed in this project, each grounded in a specific failure mode we encountered. We organize them around the failure mode and the supervision practice that emerged in response; we do not claim these are universal principles.

\subsection{Managing Autonomy in Scientific Software}
\label{sec:managing}

\paragraph{P1: Oracle testing verifies what, not why.}

An oracle test compares numerical output at fixed input parameters. It answers the question ``does the code produce the right numbers here?'' but not ``does the code produce the right numbers \emph{for the right reasons}?'' The fudge factor $\alpha = 0.27$ passed every oracle test (nine spectra, hundreds of $k$-modes, relative errors below 1\%) because it was calibrated to the same fiducial cosmology used for testing. The oracle provides no signal that the underlying model is wrong when the calibration point happens to coincide with the test point.

Multi-cosmology testing was active from project inception, but it was not what caught $\alpha=0.27$. The catch came from a limiting-case stress test: setting the tuned coefficient to $\alpha=0$ and observing that real-space errors exceeded 1\%, which exposed the calibration as compensating for a structural defect rather than encoding a physical correction. The fix was a re-derivation from the \classpt source code, implementing anisotropic damping $\Sigma^2_\mathrm{tot}(\mu)$ inside the existing Gauss--Legendre loop and dropping $\alpha$ entirely (\cref{sec:wall}). The short interval between the fudge-factor commit and its replacement shows that the supervision practice---operationalizing ``no fudge factors'' into a parameter-boundary probe---worked in this instance, but as an ad-hoc check rather than as an automated pre-commit gate. Two complementary defenses follow: test at diverse parameter points beyond the fiducial calibration so single-point calibrations are exposed when parameters shift, and automate limiting-case probes (set each tuned coefficient to a boundary value, re-run the oracle) as a mandatory pre-commit step so the fudge-rejection mechanism does not depend on a supervisor noticing.

\paragraph{P2: Shared memory prevents re-exploration but not architectural loops.}

The CHANGELOG protocol successfully prevented the agent from re-exploring solved bugs across sessions. When the DST $k$-grid issue (logarithmic vs.\ linear spacing) was resolved on day 3, no subsequent session revisited it---the CHANGELOG entry served as shared memory. However, the same protocol did not prevent 33 sessions of coherent but futile work within the wrong architecture. Sessions during the RSD crisis logged their attempts faithfully, showing no contradiction or repetition---only incremental, methodical exploration within a space that contained no solution.

The shared memory prevented \emph{literal} re-exploration (trying the same coefficient twice) but could not surface \emph{structural} stagnation (trying different coefficients within the same doomed architecture). The defense we adopted was a session-count trigger. When progress stalls for approximately 5--10 sessions with no monotonic improvement in the target metric, escalate to human review. This heuristic would have caught the RSD crisis at session 30 rather than session 56.

\paragraph{P3: The irreducible human role is architectural and physical judgment.}

The agent excelled at tasks with clear specifications and verifiable outputs: transcribing equations from a reference source, comparing intermediate numerical values to localize discrepancies, adjusting coefficients to match a known target, implementing a well-specified algorithm. These tasks, which constituted the bulk of the project's session-time, the agent performed autonomously and correctly (10 of 15 documented issues; see Appendix~\ref{app:bug-table}).

The agent failed at two specific tasks. First, recognizing when a code architecture is fundamentally incompatible with the target physics rather than merely mis-parameterized. This requires understanding what the physics \emph{should look like}---not just what numbers it should produce, but what structural properties must hold (isotropy vs.\ anisotropy, polynomial vs.\ exponential dependence, which variables are coupled). Second, detecting physically unmotivated patches that happen to improve numerical agreement. This requires knowing what the theory \emph{allows}---that a scalar correction $\alpha$ must derive from some identifiable physical mechanism, and that its absence from the reference implementation is evidence against its validity rather than evidence of a novel improvement.

Both reduce to a single underlying skill that we term \emph{explanatory agency}: evaluating explanations, not just predictions. The agent evaluates whether the output matches the target. The physicist evaluates whether the mechanism producing that output is physically coherent. When these two criteria diverge---when a wrong mechanism produces right numbers---only the physicist can detect the error.

Two interventions, complementary to scaling, could narrow the gap between autonomous and human-required bugs observed here. Retrieval-augmented reasoning over the full reference codebase could in principle surface relevant physics without human guidance, though as noted above retrieval was already complete in our case. Explicit ``physics audit'' prompting---systematically asking the agent whether every tuned parameter corresponds to a physical quantity in the reference theory---would operationalize the diagnostic question that caught the fudge factor. The physicist's intervention reduced to a single query (``does $\alpha=0.27$ appear anywhere in \classpt?''), which the agent answered correctly once asked but could not generate unprompted. Embedding such queries as mandatory checkpoints after any parameter introduction would convert an ad hoc human insight into a reproducible protocol step. The open question---whether agents with greater capacity than current LLMs would spontaneously generate the diagnostic question without prompting---is the future-research direction this case study suggests but cannot itself answer.

\subsection{Credit, Attribution, and Responsibility}
\label{sec:credit}

The division of labor raises an unavoidable question about credit. The agent performed the bulk of session-time (implementation, debugging, transcription); the physicist performed a small fraction but supplied $100\%$ of the decisive architectural and physical judgments. Contribution weight should reflect \emph{irreplaceability}, not volume: the three load-bearing interventions (the architectural redesign, the rejection of the calibration patch, and the identification of the correct anisotropic formula) were not substitutable by any agent in this study, while the autonomous bug fixes could have been done by any sufficiently capable coding agent against the same oracle.

The physicist's interventions resembled advising a graduate student. What separates the agent from a graduate student in this work is not the volume of code produced but the absence of \emph{explanatory agency}: the capacity to spontaneously generate questions about whether the current solution frame is correct, and to defend the answers under scrutiny. A student who discovered $\alpha = 0.27$ would acquire discomfort with an unphysical parameter and eventually formulate the diagnostic question without prompting; the agent could not.

Until agents develop this capacity, authorship and intellectual responsibility for AI-assisted scientific software should remain with the supervising human, and AI-assisted scientific software should ship with a supervision log as provenance documentation, analogous to a lab notebook. The supervision log for this work is publicly available alongside the code (see \cref{sec:availability}).

\subsection{What This Case Study Cannot Show}
\label{sec:cannot}

This case study is bounded in four specific ways readers should hold against any generalization.

\paragraph{Counterexamples not observed.}
The supervisor adopted a hands-off approach, intervening only after multiple sessions without progress; the agent reviewed and confirmed which dead ends were its own explorations. We observed no cases in which the supervisor's intuition was wrong, misleading, or less efficient than the agent's autonomous trajectory. This is selection bias---the supervisor intervened when the agent was stuck, not when it was on-track. A symmetric study---one in which the agent's autonomous decisions are independently judged against the supervisor's interventions---would be required to characterize the relative error rates of human and agent in this domain.

\paragraph{Inference cost not recoverable.}
Inference cost over the v0.1.0 window cannot be reported. The development host was identified retrospectively through preserved development records and system logs, but the agent session transcripts had been deleted from disk by the time of audit. Session-level cost aggregation at the time of work, not after, would have made this recoverable; we flag this as a methodological gap for future case studies.

\paragraph{Retrieval-vs-agency limitation.}
The architectural diagnosis (\cref{sec:wall}) was triggered by injecting a physics concept (anisotropic damping). We did not run a controlled ablation isolating the conceptual injection from a hypothetical navigational pointer with no physics content. The agent's codebase mapping was already complete in this case (\cref{sec:bugs}), so the relevant retrieval was already done---but we cannot rule out that a sufficiently aggressive retrieval system, delivered to a different agent that had not yet built our agent's prior survey, might have surfaced the anisotropic branch unprompted. Controlled ablations across multiple stuck-agent cases---separating process scaffolding (a generic reconsider-the-architecture prompt with no physics content), conceptual injection (a physics concept with no code pointer), and retrieval augmentation (a code pointer with no physics content), with retrieval state held constant---are a natural follow-up.

\paragraph{Single case study ($N=1$).}
This is a single case study ($N=1$): one agent architecture, one domain (cosmological perturbation theory), one supervising physicist. The intervention-level classification was reviewed by the agent, which confirmed which dead ends were its own work; this is a documented second-pass check, not an independent human inter-rater study. Row-level bug data with confidence flags and a separate independent reconstruction from the development log is provided in Appendix~\ref{app:bug-table}; readers can disagree with individual rows.

\section{Related Work}
\label{sec:related}

\citet{carlini2026}~reported on building a C compiler using 16 parallel Claude agent sessions over approximately 2,000 sessions, producing a 100,000-line Rust implementation that compiles the Linux kernel. Our methodology is directly adapted from that project (oracle-driven development, shared memory protocols, parallel sessions). The key difference is that the C compiler domain requires syntactic and semantic correctness (fully captured by oracle tests, GCC output) but not \emph{physical} correctness. Indeed, the compiler setting is precisely the case where oracle testing \emph{is} sufficient---the absence of a gap between predictive and explanatory correctness is what made fully autonomous development feasible there. Our case study extends the Carlini methodology to a domain where this gap exists and is load-bearing.

\citet{villaescusa2025} describe Denario, a multi-agent system that autonomously generates scientific analyses and manuscripts in astrophysics. Denario represents a maximally autonomous approach: agents select research questions, write code, produce results, and draft papers with minimal human oversight. Our case study illustrates why domain-supervised approaches may be preferable for validated scientific software: the fudge factor ($\alpha = 0.27$) that our agent committed, which passed all automated tests, would have been published without question in a fully autonomous pipeline. The supervision protocol catches errors that autonomy cannot.

The fudge factor failure mode connects to a broader phenomenon in AI alignment: specification gaming, where an agent optimizes a proxy metric until it ceases to measure the intended objective~\citep{krakovna2020}. The $\alpha=0.27$ correction is a domain-specific instance---the agent optimized test-suite error (the proxy) at the expense of physical correctness (the true objective). Our ``no fudge factors'' rule operationalizes the general defense: constrain optimization to exclude solutions that game the metric.

Not all AI-for-science applications require this level of domain supervision. In protein structure prediction~\citep{jumper2021} and materials discovery~\citep{merchant2023}, the oracle (experimental structures or density-functional-theory calculations from first principles) captures the full correctness criterion. There is no gap between ``right numbers'' and ``right physics'' because the target \emph{is} the numbers. Our case study applies specifically to domains where correctness requires agreement with physical theory, and where the theory admits multiple numerically adequate but physically distinct implementations.

\section{Conclusion}
\label{sec:conclusion}

In this case study, the AI agent functioned as a highly capable tool---not a co-author and certainly not an independent researcher. It could implement, debug, and validate scientific software at speed and scale impractical for a solo physicist, but it lacked the capacity to spontaneously generate the meta-question of whether its solution frame was correct. The supervision practices applied here---oracle testing against an established reference, shared session memory, the explicit ``no fudge factors'' rule operationalized into a limiting-case parameter probe, and supervisor-led escalation when sessions stalled---were the mechanism that transformed raw agent output into trustworthy scientific code. Improving this protocol, rather than solely improving model capability, was the more direct path to reliability in our case; whether the same holds beyond this one project, supervisor, and domain is a question our evidence does not settle.

\section{Code and Data Availability}
\label{sec:availability}

The \claxpt source code, the full \texttt{CHANGELOG.md} development log, the per-session intervention classification underlying Appendix~\ref{app:bug-table}, and the commit history covering the v0.1.0 development window are publicly available at \href{https://github.com/MinhMPA/clax-pt}{\faGithub\ MinhMPA/clax-pt}. The benchmark validation data used in \cref{tab:accuracy} (\classpt reference power spectra at the Planck 2018 fiducial cosmology and at the cosmology variations referenced in \cref{sec:protocol}) is released alongside the code.

\section*{Acknowledgments}

I am grateful to Siddharth Mishra-Sharma for sharing about his work and for his collaboration on \clax, the building block of \claxpt. I thank Ben Horowitz and Kazuyuki Akitsu for helpful discussions, and the two anonymous referees for useful, constructive feedback. I acknowledge support from the Japan Foundation for Promotion of Astronomy Research Grant and the JSPS KAKENHI Grant Numbers 25K23373 and 26H00404. This work was supported by World Premier International Research Center Initiative (WPI Initiative), MEXT, Japan.

\bibliography{references}

@misc{carlini2026,
  author = {Carlini, Nicholas},
  title = {Building a {C} compiler with a team of parallel {Claudes}},
  howpublished = {Anthropic Engineering Blog, \url{https://www.anthropic.com/engineering/building-c-compiler}},
  year = {2026}
}

@article{villaescusa2025,
  author = {Villaescusa-Navarro, Francisco and others},
  title = {The {Denario} project: Deep knowledge {AI} agents for scientific discovery},
  journal = {arXiv preprint arXiv:2510.26887},
  year = {2025}
}

@article{chudaykin2021,
  author = {Chudaykin, Anton and Ivanov, Mikhail M. and Philcox, Oliver H. E. and Simonovi{\'c}, Marko},
  title = {Nonlinear perturbation theory extension of the Boltzmann code {CLASS}},
  journal = {Phys. Rev. D},
  volume = {103},
  pages = {023507},
  year = {2021},
  doi = {10.1103/PhysRevD.103.023507},
  note = {\url{https://github.com/Michalychforever/CLASS-PT}}
}

@article{Carrasco:2012cv,
  author = {Carrasco, John Joseph M. and Hertzberg, Mark P. and Senatore, Leonardo},
  title = {{The Effective Field Theory of Cosmological Large Scale Structures}},
  eprint = {1206.2926},
  archivePrefix = {arXiv},
  primaryClass = {astro-ph.CO},
  doi = {10.1007/JHEP09(2012)082},
  journal = {JHEP},
  volume = {09},
  pages = {082},
  year = {2012}
}

@article{Baumann:2010tm,
  author = {Baumann, Daniel and Nicolis, Alberto and Senatore, Leonardo and Zaldarriaga, Matias},
  title = {{Cosmological Non-Linearities as an Effective Fluid}},
  journal = {JCAP},
  volume = {07},
  pages = {051},
  year = {2012},
  doi = {10.1088/1475-7516/2012/07/051},
  eprint = {1004.2488},
  archivePrefix = {arXiv},
  primaryClass = {astro-ph.CO}
}

@article{Carroll:2013oxa,
  author = {Carroll, Sean M. and Leichenauer, Stefan and Pollack, Jason},
  title = {{Consistent effective theory of long-wavelength cosmological perturbations}},
  journal = {Phys. Rev. D},
  volume = {90},
  number = {2},
  pages = {023518},
  year = {2014},
  doi = {10.1103/PhysRevD.90.023518},
  eprint = {1310.2920},
  archivePrefix = {arXiv},
  primaryClass = {hep-th}
}

@article{senatore2015,
  author = {Senatore, Leonardo and Zaldarriaga, Matias},
  title = {The {IR}-resummed Effective Field Theory of Large Scale Structures},
  journal = {JCAP},
  volume = {2015},
  number = {02},
  pages = {013},
  year = {2015},
  doi = {10.1088/1475-7516/2015/02/013}
}

@article{vlah2015,
  author = {Vlah, Zvonimir and White, Martin and Aviles, Alejandro},
  title = {A {Vlasov-Poisson} approach to the large-scale structure, with applications to the {EFT}},
  journal = {JCAP},
  volume = {2015},
  number = {09},
  pages = {014},
  year = {2015},
  doi = {10.1088/1475-7516/2015/09/014}
}

@article{simonovic2018,
  author = {Simonovi{\'c}, Marko and Baldauf, Tobias and Zaldarriaga, Matias and Carrasco, John Joseph and Kollmeier, Juna A.},
  title = {Cosmological perturbation theory using the {FFTLog}: formalism and connection to {QFT} loop integrals},
  journal = {JCAP},
  volume = {04},
  pages = {030},
  year = {2018},
  doi = {10.1088/1475-7516/2018/04/030}
}

@article{fang2017,
  author = {Fang, Xiao and Blazek, Jonathan A. and McEwen, Joseph E. and Hirata, Christopher M.},
  title = {{FAST-PT II}: an algorithm to calculate convolution integrals of general tensor quantities in cosmological perturbation theory},
  journal = {JCAP},
  year = {2017},
  volume = {2017},
  number = {02},
  pages = {030}
}

@article{Ivanov:2019pdj,
  author = {Ivanov, Mikhail M. and Simonovi{\'c}, Marko and Zaldarriaga, Matias},
  title = {{Cosmological Parameters from the BOSS Galaxy Power Spectrum}},
  eprint = {1909.05277},
  archivePrefix = {arXiv},
  primaryClass = {astro-ph.CO},
  reportNumber = {INR-TH-2019-016, CERN-TH-2019-132},
  doi = {10.1088/1475-7516/2020/05/042},
  journal = {JCAP},
  volume = {05},
  pages = {042},
  year = {2020}
}

@article{DAmico:2019fhj,
  author = {D'Amico, Guido and Gleyzes, J{\'e}r{\^o}me and Kokron, Nickolas and Markovic, Katarina and Senatore, Leonardo and Zhang, Pierre and Beutler, Florian and Gil-Mar{\'\i}n, H{\'e}ctor},
  title = {{The Cosmological Analysis of the SDSS/BOSS data from the Effective Field Theory of Large-Scale Structure}},
  eprint = {1909.05271},
  archivePrefix = {arXiv},
  primaryClass = {astro-ph.CO},
  doi = {10.1088/1475-7516/2020/05/005},
  journal = {JCAP},
  volume = {05},
  pages = {005},
  year = {2020}
}

@misc{krakovna2020,
  author = {Krakovna, Victoria and Uesato, Jonathan and Mikulik, Vladimir and Rahtz, Matthew and Everitt, Tom and Kumar, Ramana and Koop, Zac and Lefrancq, Adri\`{a} and Legg, Shane},
  title = {Specification gaming: the flip side of {AI} ingenuity},
  howpublished = {DeepMind Blog},
  year = {2020}
}

@article{jumper2021,
  author = {Jumper, John and others},
  title = {Highly accurate protein structure prediction with {AlphaFold}},
  journal = {Nature},
  volume = {596},
  pages = {583--589},
  year = {2021},
  doi = {10.1038/s41586-021-03819-2}
}

@article{merchant2023,
  author = {Merchant, Amil and others},
  title = {Scaling deep learning for materials discovery},
  journal = {Nature},
  volume = {624},
  pages = {80--85},
  year = {2023},
  doi = {10.1038/s41586-023-06735-9}
}
\bibliographystyle{icml2026}

\appendix

\section{Issue-Level Classification}
\label{app:bug-table}

This appendix lists the 15 supervision events documented during the \claxpt v0.1.0 development window. Each row is one event; the count itself involves a judgment call (a defensible range of 13--15 depending on whether a test-metric correction and the architectural redesign are counted separately---see scope note below). The classification was reviewed by a second pass that independently reconstructed the table from the same development log without seeing the original classification; rows where the two passes agreed are marked confidence~$=$~high, rows where intervention-level attribution required interpretation are marked medium or low. The development log records technical fixes but does not record which party (physicist or agent) proposed each fix; intervention-level attribution therefore rests on the supervisor's recall, cross-checked against the agent's own session-log review.

\paragraph{Scope note.}
The following items are intentionally excluded from the count: (i) test-script typos that affected test runs but not the implementation module; (ii) caveats acknowledged in the development log but never escalated to bug status (e.g., a hardcoded fallback value for the sound horizon resolved after v0.1.0; the $\sigma_v^2$ FFTLog-grid integration with $\sim$0.1\% accuracy cost); (iii) post-v0.1.0 fixes (a numerical integration API update; AD-correctness gradient fixes that landed after the v0.1.0 milestone). The 12-day window closes at the v0.1.0 milestone; later work on \claxpt extends the public record but is not part of this case study's evidence base.

\begin{table*}[h]
\caption{Issue-level classification for \claxpt v0.1.0 development. Categories: convention/unit (CV), algorithm transcription (AT), numerical coefficient (NC), architectural mismatch (AM), calibration patch (CP), test methodology (TM). Intervention levels: autonomous (A), human-accelerated (HA, physicist supplied a magnitude/shape observation), reconsider-prompt-failed-then-domain-hint (RP--DH), physicist-rejected-agent-solution (PR). Confidence is high (H) where development-log evidence directly supports the assignment, medium (M) where category is clear but intervention required interpretation, low (L) where both passes disagreed or evidence is indirect.}
\label{tab:bug-table}
\centering
\footnotesize
\setlength{\tabcolsep}{4pt}
\begin{tabular}{@{}rlllccl@{}}
\toprule
\# & Issue & Discovered & Category & Intervention & Confidence & Log ref. \\
\midrule
1  & FFTLog $M_{22}$ Hermitian vs symmetric packing      & 2026-04-02 & AT & A      & M & L218 \\
2  & $M_{22}$ row-major vs LAPACK column-major           & 2026-04-02 & CV & A      & M & L219 \\
3  & IR resummation log $k$-grid                         & 2026-04-03 & AT & A      & M & L220 \\
4  & IR resummation linear interpolation                 & 2026-04-03 & AT & A      & M & L221 \\
5  & Spurious $h^3$ multiply in bias/multipole fns       & 2026-04-04 & CV & A      & M & L274 \\
6  & $b_4$ $k$-factor $(k_h/h)^2 \to k_h^2$              & 2026-04-04 & CV & HA     & M & L275 \\
7  & Incomplete $M_{22}$ RSD kernels                     & 2026-04-04 & AT & A      & M & L276 \\
8  & Incomplete $M_{13}$ RSD kernels                     & 2026-04-04 & AT & A      & M & L277 \\
9  & UV counterterm coefficients ($\ell=2,4$)            & 2026-04-04 & NC & HA     & M & L278 \\
10 & $P_{gg,\ell=2}$ tree used isotropic component       & 2026-04-09 & AT & A      & M & L264 \\
11 & $P_{gg,\ell=4}$ tree had spurious $b_1$ factors     & 2026-04-09 & AT & A      & M & L265 \\
12 & Hexadecapole zero-crossing test metric              & 2026-04-09 & TM & A      & M & L266 \\
13 & $\alpha=0.27$ calibration patch (introduction)      & 2026-04-09 & CP & A      & H & L267 \\
14 & $\alpha$ rejection and anisotropic re-derivation    & 2026-04-09 & AM & PR     & H & L268 \\
15 & RSD architectural redesign (GL $P(k,\mu)$ assembly) & 2026-04-08 & AM & RP--DH & M & L294--L336 \\
\bottomrule
\end{tabular}
\end{table*}

The two classification passes agreed on category for all 15 rows and on intervention level for 12 rows; the three rows with intervention-level disagreement (\#6 between A and HA; \#9 between A and HA; \#15 between RP--DH and DH-only) are marked confidence M to flag this. Bug \#12 is classified as a test-methodology change rather than an implementation defect; readers preferring a stricter count may exclude it, yielding 14 rows. Bugs \#13 and \#14 represent the same parameter ($\alpha$) added and then removed within a single session 50 minutes apart; we count them as separate events because the fix was a structural re-derivation, not a numerical adjustment. Reasonable taxonomies could collapse these into one row, yielding 13 rows. The text of \cref{sec:bugs} and the abstract report the upper bound (15); readers preferring a stricter count of 13 or 14 should reach equivalent conclusions about the qualitative pattern.

Across the 15 events the intervention-level distribution is: 10 autonomous (A); 2 human-accelerated (HA); 1 architectural with explicit reconsider-prompt-failed-then-domain-hint (\#15); 1 physicist-rejected-agent-solution (\#14); 1 test-methodology correction (\#12). Session-time was distributed unevenly: rows 1--12 collectively consumed substantially fewer sessions than rows 13--15 alone, which together spanned the architectural-wall window described in \cref{sec:wall}.

\end{document}